\documentclass[10pt]{article} 
\usepackage{acl-style-files/latex/acl}

\usepackage{hyperref}
\usepackage{url}
\usepackage{xurl}
\usepackage{latexsym,amsmath}
\usepackage{booktabs}
\usepackage{graphicx}
\usepackage{subcaption}
\usepackage[nameinlink]{cleveref}
\usepackage{enumitem}
\usepackage{xcolor}
\usepackage{colortbl}
\usepackage{pgfplots}
\usepackage{pgfplotstable}
\usetikzlibrary{shapes,arrows,shadows.blur,calc,positioning,fit}

\pgfplotsset{compat=1.18}

\usepackage{times}
\usepackage[T1]{fontenc}
\usepackage[utf8]{inputenc}
\usepackage{microtype}
\usepackage{enumitem}
\usepackage{multirow}

\title{XferBench: a Data-Driven Benchmark for Emergent Language}

\author{Brendon Boldt, David Mortensen \\
Language Technologies Institute \\
Carnegie Mellon University\\
Pittsburgh, PA 15213, USA \\
\texttt{\{bboldt,dmortens\}@cs.cmu.edu} \\
}

\newcounter{comment}

\DeclareMathOperator*\mean{\text{mean}}
\DeclareMathOperator*\stdev{\text{stdev}}

\pgfplotsset{
  discard if not/.style 2 args={
    x filter/.code={
      \edef\tempa{\thisrow{#1}}
      \edef\tempb{#2}
      \ifx\tempa\tempb
      \else
        \def\pgfmathresult{inf}
      \fi
    }
  }
}

\definecolor{BestColor}{rgb}{0.5, 0.7, 1.0}
\definecolor{WorstColor}{rgb}{1.0, 0.7, 0.5}

\newcommand{\gradientcell}[3]{
    \pgfmathparse{int(round(100*(#3-#1)/(#2-#1)))))}
      \xdef\tempa{\pgfmathresult}
      \cellcolor{BestColor!\tempa!WorstColor}
 }

\tikzset{/num grad/.style n args={3}{
  column name=#1,
  column type=r, fixed, fixed zerofill, precision=2,
  postproc cell content/.append style={
    /pgfplots/table/@cell content/.add={\gradientcell{#2}{#3}{##1}}{},
  },
}}

\begin{document}

\maketitle

\begin{abstract}
In this paper, we introduce a benchmark for evaluating the overall quality of emergent languages using data-driven methods.
Specifically, we interpret the notion of the ``quality'' of an emergent language as its similarity to human language within a deep learning framework.
We measure this by using the emergent language as pretraining data for a downstream NLP tasks in human language---the better the downstream performance, the better the emergent language.
We implement this benchmark as an easy-to-use Python package that only requires a text file of utterances from the emergent language to be evaluated.
Finally, we empirically test the benchmark's validity using human, synthetic, and emergent language baselines.
\end{abstract}

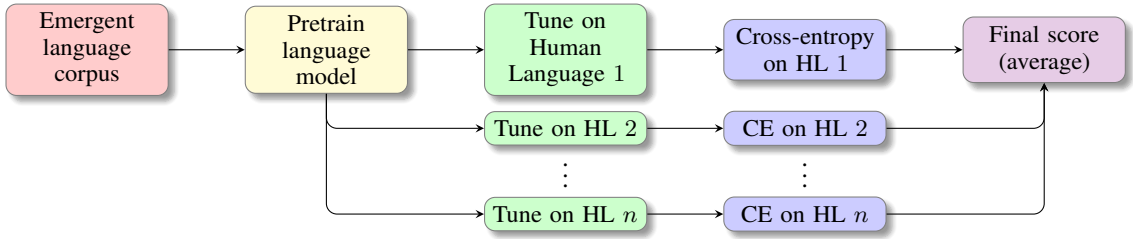
\begin{figure*}
  \centering
  \tikzstyle{blockStyle}=[
  draw=black!50,
  fill=violet!20,
  text width=1.9cm,
  minimum width=2cm,
  text centered,
  blur shadow={shadow blur steps=5},
  rounded corners,
  font=\small,
]
\tikzstyle{labelStyle} = [
  fill=black!1,
  draw=black!50,
  align=center,
  font=\small,
  text width=1.5cm,
  text centered,
  above,
  rounded corners,
]

\begin{tikzpicture}[
  node distance=10mm,
]
  \node (alpha) [blockStyle,fill=red!20] {Emergent language corpus};
  \node (beta) [blockStyle,fill=yellow!20,right=of alpha] {Pretrain language model};
  \node (gamma) [blockStyle,fill=green!20,right=of beta] {Tune on Human Language $1$};
  \node (delta) [blockStyle,fill=blue!20,right=of gamma] {Cross-entropy on HL $1$};

  \node (gamma2) [blockStyle,fill=green!20,node distance=2mm,below=of gamma] {Tune on HL $2$};
  \node (gamma3) [node distance=-1mm,below=of gamma2] {\vdots};
  \node (gamma4) [blockStyle,fill=green!20,node distance=0mm,below=of gamma3] {Tune on HL $n$};
  \node (delta2) [blockStyle,fill=blue!20,right=of gamma2] {CE on HL $2$};
  \node (delta3) at (gamma3-|delta2) {\vdots};
  \node (delta4) [blockStyle,fill=blue!20,right=of gamma4] {CE on HL $n$};
  \node (epsilon) [blockStyle,fill=violet!20,right=of delta] {Final score (average)};

  \draw [-stealth] (alpha) to (beta);
  \draw [-stealth] (beta) to (gamma);
  \draw [-stealth] (gamma) to (delta);
  \draw [-stealth] (gamma2) to (delta2);
  \draw [-stealth] (gamma4) to (delta4);
  \draw [-stealth, rounded corners] (beta.south) to (beta|-gamma2) to (gamma2.west);
  \draw [-stealth, rounded corners] (beta.south) to (beta|-gamma4) to (gamma4.west);
  \draw [-stealth] (delta) to (epsilon);
  \draw [-stealth, rounded corners] (delta2.east) to (epsilon|-delta2) to (epsilon.south);
  \draw [-stealth, rounded corners] (delta4.east) to (epsilon|-delta4) to (epsilon.south);
\end{tikzpicture}
  \caption{Illustration of the architecture of XferBench.}
  \unskip\label{fig:chart}
\end{figure*}

\section{Introduction}


Neural language models learn many things in pretraining, but research suggests \citep{artetxe-etal-2020-cross} that a substantial part of that knowledge is not simply knowledge of a particular language or domain, but rather knowledge of ``how to language.''
We currently teach models to ``language'' using vast quantities of text dredged from the dark recesses of the Web---text that is full of bias, toxicity, and potential intellectual property violations.
Ideally, we would be able to teach models to ``language'' without such compromises through the use of synthetic data, but mainstream approaches to synthesizing data produce outputs that do not have the same structural and social properties as human language.

Emergent communication (EC), also called emergent language (EL), is a potential solution to this problem \citep{yao2022linking,downey-etal-2023-learning,mu2023ec2}.
Emergent languages are communication systems developed \textit{de novo} among multiple agents in a reinforcement learning simulation.
Because the conditions under which they develop mirror, reductively, the conditions under which languages develop among humans, there is reason to believe that ELs will ultimately be more like human language than other sources of synthetic data.
However, up to this point, there is no way of quantifying---in a holistic way---how much like human languages any particular EL really is, or to what extent it may provide useful pretraining signals.

Research on deep learning-based emergent communication has seen the introduction of many metrics to measure various aspects of the language.
These metrics quantify notions such as compositionality \citep{Brighton2006UnderstandingLE,Lazaridou2018EmergenceOL}, expressivity \citep{guo2023emergent}, ease-of-teaching \citep{li2019ease}, and zero-shot transfer \citep{Bullard2020ExploringZE}, to name a few.
Despite this proliferation of metrics, emergent language largely lacks \emph{evaluation} metrics.
An evaluation metric is specifically one that measures the \emph{overall quality of an emergent language} and not simply a particular property.
Thus, we introduce XferBench, a data-driven benchmark for evaluating the overall quality of emergent languages using transfer learning with deep neural models.

Evaluation metrics are critical in gauging progress in technical fields since they quantify otherwise vague notions of improvement over time.
Benchmarks, in particular, pair evaluation metrics with specific data and evaluation procedures to compare various systems on common ground.
Benchmarks and shared tasks have been critical to the development of NLP from the Penn Treebank \citep{Marcus1993BuildingAL} to the WMT datasets \citep{bojar2014wmt} to GLUE \citep{Wang2018GLUEAM}.

In the field of emergent communication specifically,
  \citet{yao2022linking} introduced the idea of using \emph{corpus transfer} as means of practically applying emergent communication to deep learning-based NLP via transfer learning.
In corpus transfer, a language model is pretrained on a corpus of emergent language utterances before being tuned on real data for a human language-based downstream task.
As a corollary, they suggest that the effectiveness of this transfer can serve as a means of evaluating the quality of the emergent in a more general sense.
This is based on the intuition that the more similar two language are, the better transfer learning works from one to the other (observed in \citet{zoph-etal-2016-transfer}, for example).

This paper takes the transfer learning-as-an-evaluation metric idea from \citet{yao2022linking} and expands it into a full benchmark, XferBench, for emergent languages (illustrated in \Cref{fig:chart}).
An evaluation metric for emergent languages in a benchmark format is the first of its kind.
Additionally, XferBench is unique within emergent communication for being primarily data-driven instead of relying on particular handcrafted algorithms for quantifying a given phenomenon.
This means that XferBench can be easily scaled up in the future as the field of emergent communication advances and requires expanded means of evaluating emergent languages.
Finally, XferBench is distributed as a user-friendly Python package, allowing researchers from across the field of emergent communication to apply XferBench to their own work on emergent communication.


\paragraph{Contributions}
This paper makes the following contributions:
(1) Introduces XferBench, a data-driven benchmark for evaluating the overall quality of an emergent language, the first of its kind in emergent communication.
(2) Provides a analysis of the quality human, synthetic, and emergent language according to XferBench.
(3) Provides an easy-to-use Python implementation of XferBench.

\section{Related Work}


\paragraph{Emergent Communication}
This paper is situated in the field of emergent communication (a.k.a.\@ emergent language) which is generally covered by the review \citet{Lazaridou2020EmergentMC}.
The field centers around the invention of language by deep neural networks typically using multi-agent reinforcement learning techniques.
The study of emergent communication is intended to (1) shed light on the origin and nature of the human language \citep{lacroix2019biology,MoulinFrier2020MultiAgentRL,Galke2022EmergentCF} and (2) provide an alternative approach to problems in NLP and multi-agent reinforcement learning which relies on constructing language from the ground up and not just pre-existing (human) languages alone \citep{li-etal-2020-emergent,yao2022linking,mu2023ec2,downey-etal-2023-learning}.

\paragraph{Transfer Learning}
Transfer learning for deep neural networks is a key component of XferBench and follows in general tradition of \citet{zoph-etal-2016-transfer}.
Specifically, this paper draws heavily from \citet{yao2022linking} (see also \citet{Papadimitriou2020LearningMH,artetxe-etal-2020-cross}) which introduce the technique of \emph{corpus transfer} for emergent language, that is, pretraining a neural model on an emergent language corpus before tuning it on a downstream human language task.
In particular, this paper takes \citet{yao2022linking}'s idea of using corpus transfer as a metric and adapts it into a benchmark pipeline which can easily be applied to new emergent languages.

\paragraph{Benchmarks}
Work such as \citet{guo2023emergent} and \citet{perkins2022icy} have looked at benchmarking particular aspects of emergent languages, but XferBench is the first of its kind in benchmarking the overall quality of an emergent language.
\citet{yao2022linking} also explicitly provide a metric for emergent language quality, but this metric is restrictive in that it can only be applied to emergent languages derived from a model that takes images (that have captions available) as input; this conflicts with the design goals of XferBench discussed below.

Outside of emergent communication, XferBench is more analogous to benchmarks for generative models (e.g., Fr\'echet Inception Distance \citep{heusel2017fid} for image generation) than more traditional NLP benchmarks like GLUE \citep{Wang2018GLUEAM} or SQuAD \citep{rajpurkar-etal-2016-squad}.
This is because emergent communication is a generative enterprise, where one of the main goals is to create samples (emergent languages) which resemble a target distribution (human languages) either generally or in some particular respect.
Furthermore, metrics like FID are primarily self-supervised, data-driven measures of similarity in the same vein as XferBench.
This is in contrast to more traditional NLP benchmarks which combine data-driven methods with many human judgments (i.e., through labeled examples).


\section{XferBench}


\subsection{Design Goals}
\unskip\label{sec:design}

We frame the primary design goals of the benchmark as three desiderata:
\begin{enumerate}[nosep]
  \item[\textbf{D1}] Quantitatively capture a meaningful notion of the overall quality\footnotemark{} of an emergent language from a data-driven perspective.
  \footnotetext{We are aiming for \emph{a} meaningful notion of overall quality: we are not claiming that this is the only meaningful notion nor that it is the best among all possible notions of ``quality''.}
  \item[\textbf{D2}] Be applicable to as wide a variety of emergent languages as possible, not restricted to a specific game, environment, or agent architecture.
  \item[\textbf{D3}] Be relevant and accessible to the broader EC/EL community, by being:
    (a) easy to interpret,
    (b) minimally biased with regards to language typology,
    (c) runnable with minimal coding experience,
    and (d) runnable on modest hardware.
\end{enumerate}
While there are other consideration in the benchmark, these form the bulk of the motivation.
In the following paragraphs we expand upon the motivation for each design goal.

\paragraph{D1: Quantifying quality}
D1 is the core of what a benchmark seeks to do: to quantify a desirable property of a given system such that it can be compared directly to other systems (i.e., be an \emph{evaluation} metric).
There are two distinct senses in which XferBench strives towards this goal.
First, XferBench measures how good an emergent language is from a specifically machine learning perspective;
  that is, it addresses the question, ``How useful would this emergent language be for practical machine learning tasks?''
The second sense is more general: XferBench addresses the question, ``How similar is an emergent language to human language according to how deep neural networks process language?''
That is, it uses data-driven techniques to quantify the similarity between emergent language and human language in some general sense.

\paragraph{D2: Wide applicability}
D2 is intended to make XferBench practically applicable to a wide range of EC research.
The field of EC has an especially diverse set of possible approaches, environments, agents, games, etc.
Thus, it is especially salient that the benchmark be designed with interoperability in mind, having minimal assumptions as to the nature of the EC system being evaluated.

The influence of this design goal is primarily seen through the use of a textual corpus as the sole input to the benchmark: the vast majority of EC systems generate utterances which can be represented as sequences of discrete tokens.
\unskip\footnote{In the minority case, there are EC methods which use communication channels that are, for example, continuous \citep{Eloff2021TowardsLT} or even pictorial \citep{Mihai2021LearningTD}.}
EC presents the opportunity for much richer representations of its language: leveraging the grounded semantics of the communication, incorporating non-verbal behavior, and even directly interacting with the agents themselves.
Yet such richer representations also limit the range of EC systems to which XferBench could apply.
Even if it is possible to define some universal EC interface that could allow for richer representations, the implementation cost for each and every EC system to be tested is significant compared to the ease of producing a corpus of utterances from the emergent language.

\paragraph{D3: Easy-to-use}
D3 is critical to the success of XferBench as a practical tool for diverse field of researchers---a benchmark is expressly \emph{for} the broader research community, and, as such, should be widely accessible.
In particular, D3a demands that XferBench be conceptually simple with results that can easily be reported, compared, and incorporated into a research program.
D3b is relevant to both aspects of D1.
First, if XferBench is to gauge an EL's practical use in machine learning, it should seek to use a typologically diverse set of human languages in the downstream tasks.
  Second, since XferBench is trying to capture a notion of ``similarity to human language generally'', it is important to test this against a wide range of language typologies so as not to unnecessarily narrow the criteria for ``similar to human language''.
D3c is particularly important for incorporating interdisciplinary researchers into the field of EC who might not have a background in computer programming.
Finally, D3d ensures that XferBench is accessible not only to labs and researchers with fewer financial resources
  but also makes it much easier to incorporate into the fast-paced research and development cycles prevalent in contemporary ML reserach.

\subsection{Methods}
\unskip\label{sec:methods}


The following procedure describes the benchmark (illustrated in \Cref{fig:chart}):
\begin{enumerate}[nosep]
  \item Initialize a causal language model.
  \item Train the model on the corpus of utterances from the EL being evaluated.
  \item Re-initialize the input and output (i.e., language modelling head) embedding layers;
    this is the \textit{base model}.
  \item For each downstream human language:
    \begin{enumerate}
      \item Train the base model on the human language data.
      \item Evaluate the cross-entropy on a held-out test set of the human language.
    \end{enumerate}
  \item Average the cross-entropies across the downstream human languages;
    this is the corpus's score on the benchmark (lower is better).
\end{enumerate}
The structure of the benchmark is derived from the \emph{corpus transfer} method presented in \citet{yao2022linking}.

\paragraph{Task}
For XferBench's evaluation task, we choose causal language modeling for a few different reasons.
In principle, language modeling is a component of a wide variety of NLP tasks, especially generative tasks;
  the prevalence of language modeling is in line with the benchmark providing a very general notion of quality that will be familiar to anyone acquainted with NLP\@.
On a practical level, language modeling is easy to acquire data for---especially helpful for evaluating against low-resource languages---and there are fewer hyperparameters and confounding variables compared to other downstream tasks like machine translation or question-answering.
The main limitation from using language modeling is that it itself is not a widespread downstream task and so cannot guarantee direct correlation with metrics on more concrete downstream tasks (e.g., accuracy on a QA task).

For the pretraining task we also use causal language modeling.
Due to requiring a wide applicability across emergent languages (Design Goal 2), we select causal language modeling for our pretraining task since it requires only a corpus without any additional annotations or stipulations.

\paragraph{Data}
The data for the transfer learning targets (viz.\@ human languages) comes from Wikipedia dumps \citep{wikidump} (under the GFDL and CC-BY-SA 3.0 License) hosted by Hugging Face\footnotemark{}.
\footnotetext{\scriptsize\url{https://huggingface.co/datasets/wikimedia/wikipedia/tree/97323c5edeffcf4bd6786b4ed0788c84abd24b03}}
This dataset provides a diverse set of languages each with sufficient amounts of data.
For our downstream human languages, we use the same $10$ languages presented in \citet{yao2022linking}, namely:
  Basque,
  Danish,
  Finnish,
  Hebrew,
  Indonesian,
  Japanese,
  Kazakh,
  Persian,
  Romanian,
  and Urdu.
Having a variety of languages reduces the likelihood that XferBench will be biased toward specific typologies of human language (Design Goal 3b).

We use $15$ and $2$ million tokens for the pretraining and fine tuning phases, respectively following \citet{yao2022linking}.
Datasets are always repeated or truncated to fit the required size so that the number of training steps stays constant.

\paragraph{Tokenization}
For tokenization we use byte pair encoding (BPE) \citep{Gage1994ANA} with a vocabulary size of $30\,000$ for all human languages.
Using BPE across all human languages is done primarily to simplify the implementation and keep tokenization methods consistent across all of the selected human languages.
Emergent languages are generally considered to be pre-tokenized since most communication channels consist of one-hot vectors;
  thus, no additional tokenization or preprocessing is applied.
\unskip\footnote{%
  Whether the tokens of an EL should be treated as words or subword units is an open question, although tokens as words is more common (but see \citet{ueda2023on} for tokens as subword units).
  Practically speaking, many emergent languages are small enough that applying a $30\,000$-item BPE model would severely reduce the corpus size.
}
  
\paragraph{Model}

For our model, we use a small configuration of GPT-2 \citep{radford2019language}, similar to that used in \citet{yao2022linking}:
  $6$ attention heads,
  $6$ layers,
  context length of $256$,
  and hidden size of $768$
  with the remainder of the model parameters being the same as the defaults in the Hugging Face Transformers implementation.
  \unskip\footnote{\scriptsize\url{https://huggingface.co/docs/transformers/v4.36.1/en/model_doc/gpt2\#transformers.GPT2Config}}
This yields $65$ million parameters in total.
We kept the model on the smaller size to better suit it for the generally small amounts of data emergent languages corpora provide as well as to be more accessible (Design Goal 3d).
Further details are listed in \Cref{sec:hparams-clm}.

\paragraph{Metric}

Given the use of language modeling for our evaluation task, we use token-level cross-entropy as the evaluation metric on the downstream task.
This is a very common metric, making the outputs easy to interpret (Design Goal 3a).
Although perplexity is more common as an evaluation of language models, the exponential nature of perplexity leads to more circuitous analyses and interpretation in our case, whereas cross-entropy is comparatively linear and additive (loosely speaking).
\unskip\footnote{For example, it would make more sense to use logarithmic scales and geometric means to average and compare perplexities, but this would just be reverting back to cross-entropy!}
For the final score of the benchmark, we take the arithmetic mean of the cross-entropy across the $10$ downstream human languages.
That is, we define the benchmark's score for a given source language $s$ as as $h_s$:
\begin{align}
  h_{s} &= \mean_{t \in T}\left( h_{s,t}\right)
  \label{eq:hs}
\end{align}
where $h_{s,t}$ is the test cross-entropy of a model trained on source language $s$ and finetuned and tested on target language $t$;
  $T$ is the set of target languages.
Since the score is based on cross-entropy, a lower score means better performance.

\subsection{Implementation}
\unskip\label{sec:implementation}

XferBench is implemented as a small Python codebase which relies primarily on Hugging Face Transformers \citep{Wolf2019HuggingFacesTS} (Apache-2.0 license) and PyTorch \citep{pytorch} (BSD-3-Clause license) libraries.
To run the benchmark, all that is required is to install the environment with either pip or conda, and run {\small\texttt{python -m xferbench path/to/corpus.jsonl}} (Design Goal 3c).
The input corpus is simply formatted as a newline-separated list of integer arrays, specifically in the JSON Lines format (see \Cref{sec:input-example} for an example); a Hugging Face dataset (backed by Apache Arrow) can also be used for larger input corpora.
The script executes all of the steps of the benchmark and yields a single floating point number which is that corpus's score on XferBench (the benchmark also saves the individual score across target languages for further analysis).
Finer-grained functionalities are available and documented in the codebase.
The benchmark takes about $5.5$ hours to run on a single NVIDIA GeForce RTX 2080 Ti:
  $90$ minutes to train the base model and $30$ minutes for tuning and testing on each of the target languages (Design Goal 3d).
Since the model is tuned independently on each target language, it is easy to parallelize this step and drastically shorten the wall-clock time of XferBench.

The implementation is available at \url{https://github.com/brendon-boldt/xferbench} under the MIT license.


\section{Experiments}


\subsection{Procedures}
\unskip\label{sec:exp-proc}

\paragraph{XferBench}
The causal language modeling experiment is simply running XferBench as described in \Cref{sec:methods} on the reference and emergent languages discussed in \Cref{sec:ref-langs,sec:em-langs}.

\paragraph{Machine translation}
The machine translation experiment is structured similarly to XferBench except with the downstream task being English-to-French translation (using the WMT 2014 dataset \citep{bojar2014wmt}).
The primary purpose of this experiment is to determine how well XferBench correlates with a more concrete downstream task (especially one that incorporates language modeling).
We choose this language pair in part to gauge the relative differences between the task languages and the baseline human languages (in contrast to XferBench which we want to be largely agnostic to human languages).
Looking at our reference human languages, we have:
  French, the target language itself;
  Spanish, closely related to French;
  Russian and Hindi, distantly related to French;
  and Chinese, Korean and Arabic, not related to French.
Instead of using a GPT-2--based model, we use a BART-based model since MT is a conditional generation task (see \Cref{sec:hparams-mt} for details).
The pretraining dataset size is increased to $100$ million due to the increased difficulty of this task compared to language modeling.
We evaluate the translation performance with chrF \citep{chrf} and BLEU \citep{bleu} using the default Hugging Face Evaluate metrics (derived from sacreBLEU \citep{post-2018-call}).
Evaluation is performed with beam sizes of $1$, $3$, and $5$, and the resulting values are averaged.

We present three settings for this experiment.
The first is \emph{Full} which tunes on $50$ million source tokens at a higher learning  rate ($1\cdot10^{-4}$ for training and $2\cdot10^{-4}$ for the AdamW optimizer \citep{adam}), which we found empirically to lead to the best performance.
The second is \emph{Frozen}, in which we use the same configuration as \emph{Full} but freeze all but the embedding layers before tuning the model for translation (as in \citet{Papadimitriou2020LearningMH,artetxe-etal-2020-cross}).
Finally, we also present \emph{Reduced} which uses a smaller tuning dataset of $10$ million tokens and lower learning learning ($2\cdot10^{-5}$);
  the lower rate helped the random baselines converge better as well as showed better distinction between languages.

\subsection{Reference languages}
\unskip\label{sec:ref-langs}

The following reference languages serve as a way to contextualize the results of XferBench as well as to validate that it is capturing some notion of the quality of the emergent languages (cf. \Cref{sec:hypotheses}).

\paragraph{Human languages}

For our baseline line human languages, we selected French, Spanish, Russian, Chinese, Korean, Arabic, and Hindi.
\unskip\footnote{The main reason for choosing the high-resource language is due to the higher data requirements of machine translation experiment discussed below.}
Like the evaluation languages, the data is derived from Wikipedia articles (same source as the target languages).


\paragraph{Synthetic languages}

For synthetic languages, we follow \citet{yao2022linking} and use ``Zipfian parentheses'' from \citet{Papadimitriou2020LearningMH}.
This synthetic dataset---referred to as \emph{Paren, real}---is hierarchically balanced ``parentheses'' where each parenthesis is the token ID sampled from the unigram distribution of a human language (hence ``Zipfian'').
This datasets mimics both the unigram distribution of a human language as well as the basic recursive hierarchical structure.
This yields a reasonably strong yet simple baseline for synthetic data.

We also test a fully synthetic dataset (\emph{Paren, synth}) which uses the same hierarchical parenthesis generation script from \citet{Papadimitriou2020LearningMH}, replacing the data-derived unigram distribution with Zipf--Mandelbrot distribution:
\begin{align}
  f(w_i) &= \frac{1}{{(i+\beta)}^\alpha}
\end{align}
where $f(w_i)$ is non-normalized probability weight of word $w$ with $1$-based index (rank) $i$, $\alpha=1$, $\beta=2.7$ \citep{mandelbrot1953informational,Piantadosi2014-ru}.

\paragraph{Random baselines}

We use two random baselines.
The first is simply a uniform unigram distribution across the whole vocabulary with no additional structure (referred to as \emph{Random}).
This baseline sheds light on whether the optimization itself, no matter training data, primes the network in some way for transfer learning.
The second ``random'' baseline is no pretraining at all (\emph{No pretrain}); that is, a network which has been freshly initialized at the tuning stage.
This baseline helps establish whether or not pretraining on other languages has any impact beyond tuning alone.

\subsection{Emergent languages}
\unskip\label{sec:em-langs}

We present a summary of the key hyperparameters of emergent languages in \Cref{tab:ec-specs}.
The emergent language corpora below come from reproductions from existing codebases with the exception of \citet{yao2022linking}, whose emergent language corpus is available for download.
Emergent languages which have a corpus size smaller than the required size are simply repeated and shuffled as many times as necessary so that the model receives the same number of optimization steps.

\begin{table}
  \centering
  \begin{tabular}{llrrr}
    \toprule
    Setting         & Observ. & $|V|$   & $|M|$   & $|C|$ \\
    \midrule
    Disc, small     & one-hot & $6$     & $11$    & $700$ \\
    Disc, large     & one-hot & $100$   & $31$    & $100\,\text{M}$ \\
    Recon, large    & one-hot & $100$   & $31$    & $31\,\text{M}$ \\
    Mu+, CUB        & embed   & $20$    & $10$    & $1.3\,\text{M}$ \\
    Mu+, SW         & embed   & $14$    & $7$     & $1.2\,\text{M}$ \\
    Yao+            & embed   & $4028$  & $15$    & $43\,\text{M}$ \\
    \bottomrule
  \end{tabular}
  \caption{%
    Summary of key hyperparameters in the tested emergent languages.
    Observations are either one-hot vectors or embeddings.
    $|V|$, $|M|$, and $|C|$ refer to the vocabulary, message, and corpus size respectively.
  }
  \unskip\label{tab:ec-specs}
\end{table}

\paragraph{Generic signalling game}
The first set of emergent languages we test are generic versions of the of the signalling game (reference game) as implemented in EGG \citep{kharitonov-etal-2019-egg} (MIT license).
These games use one-hot vectors to represent attribute--value observations, that is, observations are elements of the set $V^{|A|}$ where $V$ is the set of values and $|A|$ is the number of attributes.
The signalling game is one of the simplest and most used games in emergent communication research.

The first two language are \emph{Disc, small} and \emph{Disc, large} which are two configurations of the discrimination version of the signalling game.
Here, the sender makes an observation and sends a message;
  then, the receiver must select the corresponding observation from a small set of potential observations (like a multiple-choice question).
The \emph{small} configuration consists of $4$ attributes and $4$ values with a small vocabulary size and medium message length;
  this setting is intended to represent a toy environment that one might find in an emergent communication paper.
The \emph{large} configuration consists of $12$ attributes and $8$ values with a larger vocabulary and longer message length.
Both environments show $5$ distractor observations to the receiver (i.e., $6$-way multiple choice).
Both settings converge to a success rate ${>}95\%$ compared to a random baseline of ${~}17\%$.

The \emph{Recon, large} environment is based on the reconstruction version of the signalling game.
In this version, the receiver does not make any observations and instead must recreate the sender's observation based on the message alone (similar to an autoencoder).
The observation space has $8$ attributes and $8$ values with other settings identical to that of \emph{Disc, large}. 
Since the reconstruction game considerably harder, the game does not converge but does reach an overall accuracy of $0.014\%$ and per-attribute accuracy of $24\%$ compared to a random baseline of $0.000006\%$ and $13\%$ random baseline, respectively.
For details, see \Cref{sec:hparams-egg}.

\paragraph{\citet{mu2021generalizations}}
present the second pair of emergent languages which we test XferBench on (code under MIT license).
The emergent communication game is also a discriminative signalling game but with (1) richer observations and (2) more abstract information needing to be communicated.
In one setting, the observations are images from ShapeWorld \citep{Kuhnle2017ShapeWorldA} (\emph{Mu+, SW}), a synthetic data of various geometric shapes, and the other setting is CUB \citep{WahCUB_200_2011} (\emph{Mu+, CUB}) which contains labeled images of birds;
  both settings encode features with a CNN which is the passed to the sender and receiver.
In the basic discriminative game, the observation made by the sender is the exact same one seen by the receiver.
\citet{mu2021generalizations} instead uses a ``concept game'' where the sender must communicate some abstract concept shared by a set of input images which the receiver will then have to a pick out from a different set of images, some sharing the same concept (e.g., isolating the concept of ``triangle'' or ``bird size'').
The ShapeWorld and CUB games had test accuracies of $71\%$ and $66\%$ respectively compared to a random baseline of $50\%$, comparable to the reported values in the paper.
All messages were taken from observations seen in training.


\paragraph{\citet{yao2022linking}} present a standard discrimination game which uses natural images (Conceptual Captions \citep{sharma-etal-2018-conceptual} (images only)) as inputs to the sender and receiver (code unlicensed but distributed on GitHub with paper).
The accuracy for the particular emergent language corpus is not reported in the paper, but comparable experiments from the paper would suggest that it converged to an accuracy of ${>}90\%$ compared to a baseline of $0.4\%$ (i.e., $255$ distractors).

\subsection{Hypotheses}
\unskip\label{sec:hypotheses}

The following hypotheses are directly relate to determining whether or not XferBench is quantifying some meaningful notion of the quality of a language (i.e., Design Goal 1).

(H1) Human languages will perform best, followed by the synthetic and emergent languages, followed by the random baselines.

(H2) Human languages will have similar performance on XferBench (also key for Design Goal 3b);
  the intuition here is that human languages share deep structural similarities.
  This hypothesis is supported, in part, by \citet{artetxe-etal-2020-cross}.
  For the MT experiment, we expect to see the following order of performance based on language relatedness:
    \{\emph{French}\},
    \{\emph{Spanish}\},
    \{\emph{Russian}, \emph{Hindi}\},
    \{\emph{Chinese}, \emph{Korean}, \emph{Arabic}\}.

(H3) Languages with a larger vocabulary, longer message length, and larger corpora will perform better.
  In particular, we expect \emph{Disc, large} will perform better than \emph{Disc, small} since the former is a more ``complex'' version of the latter.
  This hypothesis (for vocabulary size and message length) is supported by some experiments in \citet[app.\@ B.4]{yao2022linking}.

(H4) XferBench will correlate well with scores on the machine translation task (i.e., cross-entropy will correlate negatively with chrF).

\section{Results}

\begin{figure*}
  \centering
  \def\datafile{data/ce-means.tsv}

\pgfplotstableread[col sep=tab]{\datafile}\mydata
\pgfplotstablegetrowsof{\mydata}
\edef\numberofrows{\pgfplotsretval}

\newcommand{\myaddplot}[2]{%
  \addplot [
    fill=#2,
    error bars/y dir=both,
    error bars/y explicit,
    discard if not={group}{#1},
  ] table [
    col sep=tab,
    y=mean,
    x=xpos,
    y error plus=plus,
    y error minus=minus,
  ] {\datafile};
}

\begin{tikzpicture}
\begin{axis}[
  ybar,
  enlarge x limits=0.1,
  enlarge y limits=0.1,
  height=2in,
  xlabel={Language},
  ylabel={Cross-Entropy},
  xticklabels from table={\mydata}{name},
  xtick={1,...,\numberofrows},
  x tick label style={rotate=30,anchor=east},
  /pgf/bar shift=0pt,
  width=\linewidth,
  legend style={at={(0.02, 0.95)}, anchor=north west, font=\small},
]

\myaddplot{human}{red!20}
\addlegendentry{Human}
\myaddplot{synth}{blue!20}
\addlegendentry{Synthetic}
\myaddplot{ec}{green!20}
\addlegendentry{EC}
\myaddplot{baseline}{black!20}
\addlegendentry{Baseline}

\end{axis}
\end{tikzpicture}
  \caption{%
    Average cross-entropy on target language datasets for each source language.
    Lower is better.
    Error bars represent $95\%$ confidence intervals.
  }
  \unskip\label{fig:ces}
\end{figure*}
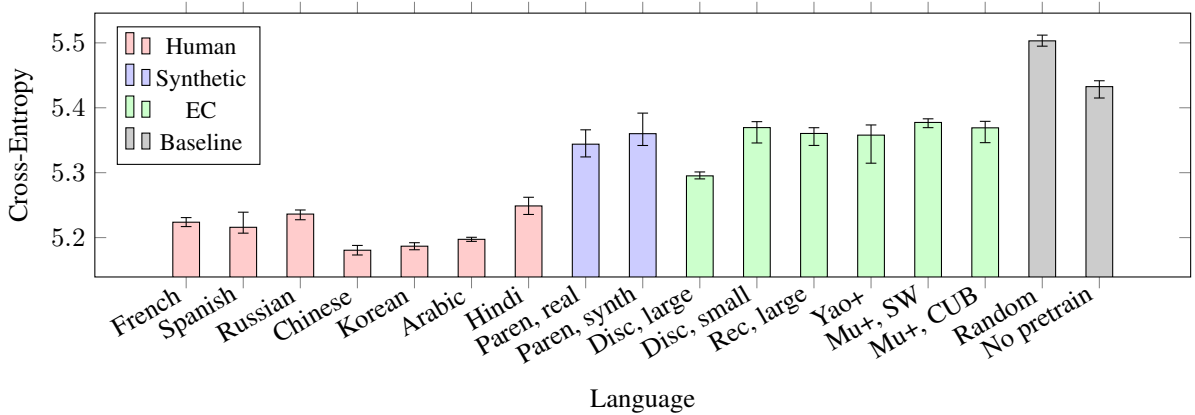

\subsection{XferBench}

In \Cref{fig:ces} we show the results of the benchmark (i.e., causal language modeling) on the various baselines.
Each mean is displayed with error bars showing the $95\%$ confidence interval of mean as calculated with bootstrapping (details in \Cref{sec:bootstrapping}).
For reference, the cross-entropies range from about $5.2$ to $5.5$ corresponding to perplexities of $180$ to $240$.

The human languages show the best score (lowest cross-entropy) on the benchmark with \emph{Chinese}, \emph{Korean}, and \emph{Arabic} performing the best in one cluster and \emph{French}, \emph{Spanish}, \emph{Russian}, and \emph{Hindi} performing slightly worse in their own cluster (based on confidence intervals).
The synthetic and emergent languages all show similar performance with only small variations with the exception of the \emph{Disc, large} language which is better than the rest of the emergent languages but still worse than the human languages.
Finally, the random baselines perform worse than the rest of the tested languages.
\emph{No pretrain}'s performance is worse than the cluster of synthetic and emergent languages but better than the fully random language (\emph{Random}).

\subsection{Machine Translation}

\begin{table}[t]
  \centering
\def\minMtFreeze{3.0008366006472933} \def\maxMtFreeze{31.43874145449612} \def\minMtLL{17.302641642978895} \def\maxMtLL{37.24975237865812} \def\minMtWmt{1.808097831638931} \def\maxMtWmt{48.238421694703696}

\begin{filecontents*}{data/mt-chrf.tsv}
key	name	xpos	MtFreeze	MtLL	MtWmt
fr	French	1	31.43874145449612	35.759689006621805	47.829632929927094
es	Spanish	2	27.947617711338808	34.793151939227464	48.04784213486238
ru	Russian	3	29.03460054134939	37.24975237865812	47.60072693560543
zh	Chinese	4	22.16511079425617	35.18695882302875	47.48953501160444
ko	Korean	5	23.3342820621439	35.62131631958034	47.726811054654114
ar	Arabic	6	27.593383406757386	36.64599210391044	47.83652459202537
hi	Hindi	7	26.03534875204922	31.66851476335667	47.496426861662506
pz_real	Paren, real	8	10.46417271167862	34.95621290271466	47.46982144423841
pz_syn	Paren, synth	9	12.048223803232311	34.26320059376425	48.238421694703696
disc_large	Disc, large	10	24.689611477604245	30.73882182351124	47.682450781549505
disc_small	Disc, small	11	16.22354976411741	17.302641642978895	14.270729332178059
recon_1	Rec, large	12	18.374409208092313	25.35174977038324	22.491934180296344
yao	Yao+	13	20.051474723736252	25.55397649295927	3.987143167253534
mu_sw	Mu+, SW	14	18.433243822247537	23.28370052105332	3.3093823147931176
mu_cub	Mu+, CUB	15	21.574161122160948	24.579589823546836	47.59033171740222
rand	Random	16	3.0008366006472933	19.666147127996197	1.808097831638931
no_pt	No pretrain	17	4.3430220150942285	28.080276800221544	11.425941817724043
\end{filecontents*}
\pgfplotstabletypeset[
  col sep=tab,
  columns={name,MtWmt,MtFreeze,MtLL},
  columns/name/.style={
    column name=Source,
    column type=l,
    string type,
  },
  columns/MtFreeze/.style={
    /num grad={Frozen}{\minMtFreeze}{\maxMtFreeze},
    precision=1,
  },
  columns/MtWmt/.style={
    /num grad={Full}{\minMtWmt}{\maxMtWmt},
    precision=1,
  },
  columns/MtLL/.style={
    /num grad={Reduced}{\minMtLL}{\maxMtLL},
    precision=1,
  },
  every head row/.style={before row=\toprule,after row=\midrule},
  every last row/.style={
    after row={%
      \midrule
      \begin{minipage}[c]{0.6in}\centering\small Correl.\@ with\\XferBench \end{minipage} & $-0.75$ & $-0.84$ & $-0.79$ \\
      \bottomrule
    }
  },
]{data/mt-chrf.tsv}

  \caption{%
    chrF scores across three English-to-French machine translation settings.
    Correlation measured with the Pearson correlation coefficient.
    Colors normalized by column.
  }
  \unskip\label{fig:mt}
\end{table}

The chrF scores of the machine translation experiment are given in \Cref{fig:mt} (BLEU scores in \Cref{sec:mt-bleu}).
Additionally, we give Pearson correlation coefficients between each setting and the scores generated by XferBench (scatter plots shown in \Cref{sec:scatter}).
In all settings, we see that XferBench is strongly correlated with the results of the machine translation experiment.

For the \emph{Full} setting, the results are somewhat inconclusive.
Human languages perform the best and similarly to each other.
\emph{Paren, real},
  \emph{Paren, syn},
  \emph{Disc, large},
  and \emph{Mu+, CUB}
  all match the performance of human languages as well.
The rest of the language perform significantly worse than the aforementioned languages, especially \emph{Yao+} and \emph{Mu+, SW} (see \Cref{sec:error-analysis} for sample outputs).
In the case of \emph{Random}, the training loss did not decrease during training likely due to the high learning rate.

In \emph{Frozen}, we see the best correlation with the hypothesis regarding human languages (as well as with XferBench itself).
\emph{Disc, large} performs comparably to the worst human languages and better than the rest of the languages.
The remainder of the synthetic and emergent languages perform worse than the human languages but better than the random baselines.

Finally, \emph{Reduced} (i.e., lower learning rate and tuning data) displays better separation than \emph{Full}, but not as significant as \emph{Frozen}.
Human languages still perform the best, although they are matched by the \emph{Paren} languages.
\emph{Disc, large} underperforms the human languages but still outperforms all other emergent languages.
All emergent languages, apart from \emph{Disc., large} underperform the \emph{No pretrain} baseline.
The better half of languages performed better (compared to themselves) with a higher learning rate while the lower half performed better with a reduced learning rate.

\section{Discussion}

\subsection{Experiments}

The basic ordering of the language by XferBench follows basic \emph{a priori} assumptions:
  random baselines perform the worst,
  human languages perform the best,
  and emergent and synthetic languages are bounded above and below by these
  (supporting Hypothesis 1).
Human languages cluster together in XferBench although there is still variation with non-overlapping confidence intervals (partially supporting Hypothesis 2).

\paragraph{Intra-EL differences}
Generally speaking, there is very little variation shown by XferBench on the emergent languages; nevertheless, we can still draw a handful of conclusions.
First, \emph{Disc, large} outperforms \emph{Disc, small} while sharing the same codebase, task, etc.\@ and differing only in message length, vocabulary size, observation space, and corpus size (supporting Hypothesis 3).
This result matches the trend seen in \citet{yao2022linking} that larger vocabularies and message lengths in an emergent language lead to better performance on downstream data.
On the other hand, \emph{Disc, small} performs similarly to other languages with larger vocabularies and longer message lengths (contradicting Hypothesis 3).

Second, it seems that the underlying complexity of the emergent communication game does not directly correlate with XferBench score: the abstract visual reasoning of \emph{Mu+, SW} and \emph{Mu+, CUB} does not lead to it outperform \emph{Disc, small}.
Additionally, the richer observations (i.e., image embeddings) of \emph{Mu+, CUB} and \emph{Yao+} also do not, by their mere presence, confer an advantage to the emergent language with respect to XferBench.

Finally, \emph{Disc, large} and \emph{Recon, large} both share hyperparameters in terms of the vocabulary size, message length, and corpus size, yet \emph{Disc, large} shows significantly better performance on XferBench.
This indicates that XferBench is not \emph{solely} concerned with surface-level features as we see that the nature of the game (e.g., discrimination versus reconstruction, success rate) is relevant as well.

\paragraph{Correlation with MT}
The results from the machine translation experiment show strong, though not perfect, (negative) correlation with XferBench (supporting Hypothesis 4).
For example, in all cases, \emph{Disc, large} outperforms all other emergent languages.
This strongly supports the notion that XferBench performance is predictive of downstream performance on more concrete NLP tasks.

The results from the \emph{Full} setting of the MT experiment do show some correlation with XferBench but fail to show expected trends in other ways.
For example, there is no clear ordering among the human languages (e.g., \emph{French} does \emph{not} outperform \emph{Arabic}).
Additionally \emph{Yao+} and \emph{Mu+, SW} drastically underperform the other emergent languages and the \emph{No pretrain} baseline.
We suspect that these aberrations from expected results come in part due to the high learning rate which cause unstable training or generation.
On the other hand, the \emph{Frozen} setting gives us the clearest ordering of human languages that matches with \emph{a priori} expectations; this setting also has the strongest correlation with XferBench scores.
The \emph{Reduced} setting shows better correlation than \emph{Full} but is not as clear as \emph{Frozen}.

\paragraph{Random baselines}
In all of our experiments, the pretraining on random tokens (\emph{Random}) performed notably worse than not pretraining at all (\emph{No pretrain}), suggesting that ill-conditioning the neural network can be a significant hindrance to performing well on XferBench.
This is important to note in light of the fact that a perfectly one-to-one compositional language describing uniformly sampled attribute--value vectors would yield a corpus with a uniformly random unigram distribution.
This is to say, a fully compositional language, which is often seen as desirable in emergent communication research, could make for a very poor source of pretraining data as shown by \emph{Random}'s performance on XferBench.

This fact along with the observations about sensitivity to learning rate indicates that performance on XferBench is not simply a function of the particular features of the emergent language in relation to the downstream human languages but also a function of the dynamics of optimization (i.e., priming the model to adapt well).
Although this increases the difficulty of developing and interpreting a tool like XferBench, it is almost an unavoidable part of deep learning methods.

\subsection{Future work}
We identify three main directions for future work with XferBench.
The first direction is determining what XferBench is measuring and how its scores correlate with the different factors of emergent languages.
\citet[app.\@ B.4]{yao2022linking} pursued this on a small scale with factors like vocabulary size and message length, but there exist a host of other factors worth exploring: speaker model size, game design, language entropy, observation modality, etc.

The second direction is more extensively investigating the correlation of XferBench with downstream tasks.
We would expect that tasks that rely heavily on a language model---such as automatic speech recognition, abstractive summarization, and generative question-answering---to correlate well with XferBench.
On the other hand, tasks that are more focused on classification---such as named entity recognition, sentiment analysis, and multiple choice question-answering---might not correlate as well.

Finally, XferBench would benefit greatly from improved compute efficiency.
For example, if the results of XferBench could be replicated with a fraction of the training steps, it could (1) allow for a larger number of downstream languages to be tested which would reduce the size of the confidence intervals, allowing more more precise scoring.
And (2), it would open the door to using larger models which would better capture the deeper structures of language and likely correlate better with realistic downstream tasks.

\section{Conclusion}
In this paper we have introduced XferBench, a first-of-its-kind benchmark for evaluating the quality of an emergent language corpus based on its transfer learning performance on human languages.
This approach to evaluating emergent language scales with data and compute as opposed to requiring increasingly complex handcrafted rules to measure the desirable qualities of emergent language.
We provide empirical results of XferBench across human, synthetic, and emergent languages and demonstrate that these results correlate with downstream performance on a machine translation task.
XferBench is implemented as an easy-to-use Python package that will permit researchers in the field to easily apply XferBench to new emergent languages.

\section{Limitations}
The first limitation of XferBench is that it relies on a restricted interface with the emergent communication system.
With emergent communication we have access not only to the grounding of all of the utterances of the emergent language but also full access to the agents themselves.
Language is fundamentally a contextual phenomenon, so only a small part of it can be understood from looking at corpora in isolation.
Thus, although XferBench is much more broadly applicable because of this restricted interface, it is also quite limited in what it can detect from a theoretical point of view.

The other set of limitations we will discuss have to do with the model and data size.
First, the model and data size ($60\,\text{M}$ parameters and $15\,\text{M}$ tokens) are quite small by contemporary standards, limiting the direct applicability of results from XferBench to relevant downstream tasks involving large language models, for example.
On the other hand, scaling up the models, data, and methods of XferBench comes with its own difficulties.
First, it would start to bias the benchmark towards high-resource languages, as only those could provide the necessary data to accommodate larger models.
Second, it would make XferBench, which is already relatively slow as a metric ($6$ GPU-hours) even slower.
This would decrease the speed of the iterative design process of emergent communication systems and, thus, the utility of the metric as a whole.


\bibstyle{acl-style-files/latex/acl_natbib}
\bibliography{src/main}

\appendix

\section{Hyperparameters}
\unskip\label{sec:hparams}

\subsection{Causal language modeling}
\unskip\label{sec:hparams-clm}

For values not listed, see Hugging Face Transformers' defaults at \url{https://huggingface.co/docs/transformers/v4.36.1/en//model_doc/gpt2\#transformers.GPT2Config}.
\begin{itemize}[itemsep=-1.2ex]
  \item Model: GPT-2
  \item Tokenizer: Byte pair encoding
  \item Hidden size: $768$ (default)
  \item Vocabulary size: $30\,000$
  \item Context length: $256$
  \item Number of layers: $6$
  \item Number of attention heads: $6$
  \item Learning rate: $1\cdot10^{-4}$
  \item Optimizer: AdamW
  \item Weight decay: $0.01$
  \item Learning rate schedule: linear (to $0$)
  \item Batch size: $32$
  \item Train dataset size: $15\cdot10^6$ tokens
  \item Train epochs: $5$
  \item Tune dataset size: $2\cdot10^6$ tokens
  \item Train epochs: $10$
\end{itemize}

\subsection{Machine translation}
\unskip\label{sec:hparams-mt}
For values not listed, see Hugging Face Transformers' defaults at \url{https://huggingface.co/docs/transformers/v4.36.1/en/model_doc/bart\#transformers.BartConfig}.
The following is for the \emph{Full} setting.
\begin{itemize}[itemsep=-1.2ex]
  \item Model: BART
  \item Training objective: text infilling only (see note below)
  \item Tokenizer: Byte pair encoding
  \item Hidden size: $512$
  \item Vocabulary size: $30\,000$
  \item Context length: $512$
  \item Number of encoder layers: $6$
  \item Number of decoder layers: $6$
  \item Number of encoder attention heads: $8$
  \item Number of decoder attention heads: $8$
  \item Encoder feedforward dimension: $2048$
  \item Decoder feedforward dimension: $2048$
  \item Train learning rate: $1\cdot10^{-4}$
  \item Tune learning rate: $2\cdot10^{-4}$
  \item Optimizer: AdamW
  \item Weight decay: $0.01$
  \item Learning rate schedule: linear (to $0$)
  \item Batch size: $32$
  \item Train dataset size: $100\cdot10^6$ tokens
  \item Train epochs: $5$
  \item Tune dataset size: $50\cdot10^6$ tokens
  \item Train epochs: $3$
  \item Test beam size: $1,3,5$ (final metric averaged across each size)
  \item Test context size: $128$
\end{itemize}
The objective used to pretrain BART was text infilling \emph{only}; we cannot use the sentence permutation objective because we do not know \emph{a priori} what constitutes a sentence in an emergent language, hence we do not use it for any settings.
For the \emph{Frozen} setting, all is as above, but all non-embedding layers are frozen for the duration of tuning.
For the \emph{Reduced} setting, all is as above except for the following:
\begin{itemize}[itemsep=-1.2ex]
  \item Tune learning rate: $1\cdot10^{-5}$
  \item Tune dataset size: $10\cdot10^6$
\end{itemize}

\subsection{Generic signalling game}
\unskip\label{sec:hparams-egg}
We use the following hyperparameters for the \emph{Disc, small} emergent language.
\begin{itemize}[itemsep=-1.2ex]
  \item Game (from EGG): \\\texttt{egg.zoo.basic\_games.play}
  \item Message optimization: Gumbel-softmax (as opposed to REINFORCE)
  \item Game type: discrimination
  \item Number of attributes: $4$
  \item Number of values: $4$
  \item Number of distractors: $5$
  \item Vocabulary size: $6$
  \item Max message length: $10$
  \item Number of examples: $32\,768$
  \item Batch size; $1024$
  \item Number of epochs: $10$
  \item Sender hidden size: $256$
  \item Receiver hidden size: $512$
  \item Sender embedding size: $32$
  \item Receiver embedding size: $32$
  \item Sender network type: GRU
  \item Receiver network type: GRU
  \item Learning rate: $0.001$
\end{itemize}
The \emph{Disc, large} setting uses the same hyperparameters as above with the exception of the following.
\begin{itemize}[itemsep=-1.2ex]
  \item Number of attributes: $12$
  \item Number of values: $8$
  \item Number of distractors: $5$
  \item Number of examples: $3.5\cdot10^6$
  \item Max message length: $30$
  \item Vocabulary size: $100$
  \item Number of epochs: $100$
\end{itemize}
The \emph{Recon, large} setting is as in \emph{Disc, large} with the following changes.
\begin{itemize}[itemsep=-1.2ex]
  \item Game type: reconstruction
  \item Number of attributes: $8$
  \item Number of distractors: N/A
  \item Number of examples: $1\cdot10^6$
  \item Number of epochs: $10$
\end{itemize}

\section{Example of benchmark input format}
\unskip\label{sec:input-example}

The input format for the benchmark is simple: integer arrays in a JSON format separated by newlines (i.e., JSON Lines, JSONL, {\small\tt{}*.jsonl}).
The following is an example of file contents in this format:
\begin{verbatim}
[3, 1, 4, 1, 5, 9, 2]
[6, 5, 3, 5, 8, 9, 7, 9, 3]
[2, 3, 8, 4]
[6, 2, 6, 4, 3, 3]
[8, 3, 2, 7, 9, 5, 0, 2, 8, 8, 4]
\end{verbatim}

\section{Computing resources used}
See \Cref{tab:compute} for rough estimates of the compute used in writing this paper.
Most experiments were run on a shared cluster comprising approximately $150$ NVIDIA A6000 (or comparable) GPUs.
\begin{table}
  \centering
  \begin{tabular}{lrrr}
    \toprule
    Item & Base GH & $n$ items & Total \\
    \midrule
    XferBench         & $6$ & $45$ & $270$ \\
    MT                & $8$ & $50$ & $400$ \\
    Other experiments & $2$ & $50$ & $100$ \\
    \midrule
    Total             & & & $770$ \\
    \bottomrule
  \end{tabular}
  \caption{Estimate of compute used for this paper in GPU-hours (specifically NVIDIA RTX 2080 Ti--hours).}
  \unskip\label{tab:compute}
\end{table}

\section{Additional results}

\subsection{BLEU scores for machine translation}
\unskip\label{sec:mt-bleu}
See \Cref{tab:mt-bleu}.
\begin{table}
  \centering
\def\minMtFreeze{0.0005419284672212063} \def\maxMtFreeze{5.332872693239323} \def\minMtLL{0.3837410752007806} \def\maxMtLL{7.021856958069873} \def\minMtWmt{0.0} \def\maxMtWmt{13.323590858337544}

\begin{filecontents*}{data/mt-bleu.tsv}
key	name	xpos	MtFreeze	MtLL	MtWmt
fr	French	1	5.332872693239323	6.608504012929764	12.929443377044677
es	Spanish	2	4.523495297217078	6.353324278301611	13.323590858337544
ru	Russian	3	4.368787663709368	7.021856958069873	12.925237547165615
zh	Chinese	4	3.0397946948608983	6.0309104590856935	12.707558887053358
ko	Korean	5	2.9548939671474415	6.356895582886708	12.825673059680872
ar	Arabic	6	4.160476937945206	6.738005393866072	13.123667506869197
hi	Hindi	7	3.203788970871212	5.243201791570945	12.723870456756101
pz_real	Paren, real	8	0.6451194343287119	6.261086868143855	12.596343315254122
pz_syn	Paren, synth	9	0.8155098897438009	6.145796864995524	13.194674904463902
disc_large	Disc, large	10	2.0805373498551	4.435146879696963	12.926100173223581
disc_small	Disc, small	11	0.190527407753092	0.3837410752007806	0.16908099603058488
recon_1	Rec, large	12	0.8596674601878694	2.5037117810428517	1.9231142837342319
yao	Yao+	13	1.0390550999365085	2.573379731464934	0.01278122106318877
mu_sw	Mu+, SW	14	1.0518470206119097	1.8572408086222822	0.0019351189705201974
mu_cub	Mu+, CUB	15	1.445422768770813	2.349862899688061	12.706866641129317
rand	Random	16	0.0005419284672212063	1.0176575308329152	0.0
no_pt	No pretrain	17	0.05995444143349249	3.4318703046295984	0.10217863128445374
\end{filecontents*}
\pgfplotstabletypeset[
  col sep=tab,
  columns={name,MtWmt,MtFreeze,MtLL},
  columns/name/.style={
    column name=Source,
    column type=l,
    string type,
  },
  columns/MtFreeze/.style={
    /num grad={Frozen}{\minMtFreeze}{\maxMtFreeze},
  },
  columns/MtWmt/.style={
    /num grad={Full}{\minMtWmt}{\maxMtWmt},
  },
  columns/MtLL/.style={
    /num grad={Reduced}{\minMtLL}{\maxMtLL},
  },
  every head row/.style={before row=\toprule,after row=\midrule},
  every last row/.style={after row=\bottomrule},
]{data/mt-bleu.tsv}

  \caption{BLEU scores for machine translation experiment.  Colors normalized by column.}
  \unskip\label{tab:mt-bleu}
\end{table}

\subsection{Raw cross-entropies on XferBench}
\unskip\label{sec:clm-all}
See \Cref{tab:clm-all}.
\begin{table*}
  \centering
  \pgfplotstableread[col sep=tab]{data/clm.tsv}{\mytable}
\pgfplotstablegetrowsof{\mytable}
\pgfmathtruncatemacro\LastRowNo{\pgfplotsretval-1}

\def\minda{4.888070106506348} \def\maxda{5.2285661697387695} \def\minfi{5.571471214294434} \def\maxfi{5.922538757324219} \def\minhe{5.425610065460205} \def\maxhe{5.7144880294799805} \def\minro{5.118478298187256} \def\maxro{5.449519157409668} \def\minfa{5.008924961090088} \def\maxfa{5.305185317993164} \def\mineu{6.005818843841553} \def\maxeu{6.167392253875732} \def\minja{5.180477142333984} \def\maxja{5.550778388977051} \def\minur{4.381984710693359} \def\maxur{4.718813419342041} \def\minkk{5.438832759857178} \def\maxkk{5.755498886108398} \def\minid{4.756775856018066} \def\maxid{5.218039512634277} \def\minmean{5.180562829971313} \def\maxmean{5.50308198928833}

{
  \footnotesize
  \begin{filecontents*}{data/clm.tsv}
	da	fi	he	ro	fa	eu	ja	ur	kk	id	name	group	xpos	mean
fr	4.927217483520508	5.6191864013671875	5.480484962463379	5.148122787475586	5.042868137359619	6.032809734344482	5.22976541519165	4.427328586578369	5.4606523513793945	4.868746280670166	French	human	1	5.223718214035034
es	4.918011665344238	5.612151622772217	5.469159126281738	5.124547958374023	5.033268451690674	6.056143283843994	5.245471477508545	4.423867225646973	5.4550018310546875	4.821191787719727	Spanish	human	2	5.215881443023681
ru	4.942409515380859	5.646069049835205	5.4823760986328125	5.13839054107666	5.038311004638672	6.041613578796387	5.265775680541992	4.449263572692871	5.4818854331970215	4.876440048217773	Russian	human	3	5.236253452301026
zh	4.889410495758057	5.577518463134766	5.425610065460205	5.118478298187256	5.008924961090088	6.020379066467285	5.180477142333984	4.38922119140625	5.438832759857178	4.756775856018066	Chinese	human	4	5.180562829971313
ko	4.888070106506348	5.571471214294434	5.440714359283447	5.1215667724609375	5.0174994468688965	6.005818843841553	5.2042012214660645	4.381984710693359	5.452763557434082	4.783656597137451	Korean	human	5	5.186774682998657
ar	4.896095275878906	5.591085910797119	5.45055627822876	5.127626419067383	5.02038049697876	6.0205559730529785	5.220425128936768	4.39775276184082	5.442032337188721	4.806921005249023	Arabic	human	6	5.197343158721924
hi	4.943386077880859	5.654071807861328	5.467916965484619	5.199193000793457	5.076722621917725	6.061072826385498	5.286042213439941	4.4571709632873535	5.5154571533203125	4.826949596405029	Hindi	human	7	5.2487983226776125
pz_real	5.068382263183594	5.752814292907715	5.593005657196045	5.223895072937012	5.114136219024658	6.114029407501221	5.377829074859619	4.562935829162598	5.569694519042969	5.062579154968262	Paren, real	synth	8	5.3439301490783695
pz_syn	5.075646877288818	5.738699436187744	5.57607364654541	5.264694690704346	5.142731666564941	6.132684707641602	5.428257465362549	4.569953918457031	5.578944683074951	5.093385219573975	Paren, synth	synth	9	5.3601072311401365
disc_large	5.000357627868652	5.705007076263428	5.521820545196533	5.246131420135498	5.105435371398926	6.06104850769043	5.337141513824463	4.493499755859375	5.556825637817383	4.9238762855529785	Disc, large	ec	10	5.295114374160766
disc_small	5.085509777069092	5.7980170249938965	5.589393615722656	5.31077241897583	5.174932956695557	6.0629448890686035	5.412372589111328	4.563511848449707	5.645544528961182	5.051603317260742	Disc, small	ec	11	5.3694602966308596
recon_1	5.086837291717529	5.788028717041016	5.568704128265381	5.299289703369141	5.157498359680176	6.064957141876221	5.410421371459961	4.554931640625	5.635402679443359	5.038140773773193	Rec, large	ec	12	5.360421180725098
yao	5.071822643280029	5.793422222137451	5.564732074737549	5.305386543273926	5.166547775268555	6.033895015716553	5.408265113830566	4.560029983520508	5.646118640899658	5.028499603271484	Yao+	ec	13	5.357871961593628
mu_sw	5.089369773864746	5.798652172088623	5.577664852142334	5.325026035308838	5.184543132781982	6.104068279266357	5.422131538391113	4.579131603240967	5.646788597106934	5.046632289886475	Mu+, SW	ec	14	5.377400827407837
mu_cub	5.079919815063477	5.789093017578125	5.58148717880249	5.315103530883789	5.178103446960449	6.063284397125244	5.416987419128418	4.564374923706055	5.653202533721924	5.049468994140625	Mu+, CUB	ec	15	5.369102525711059
rand	5.2285661697387695	5.922538757324219	5.7144880294799805	5.449519157409668	5.305185317993164	6.167392253875732	5.550778388977051	4.718813419342041	5.755498886108398	5.218039512634277	Random	baseline	16	5.50308198928833
no_pt	5.167593002319336	5.846884727478027	5.660545349121094	5.377307891845703	5.234503269195557	6.098935127258301	5.472503185272217	4.651296615600586	5.675395488739014	5.140706539154053	No pretrain	baseline	17	5.432567119598389
mean	5.0210944624508125	5.717924230238971	5.539101937237908	5.240885426016415	5.117740743300494	6.067154884338379	5.3452262317433075	4.514415797065286	5.565296565785127	4.9643301683313705	\textit{Mean}		18	5.309317044650808
\end{filecontents*}
\pgfplotstabletypeset[
    col sep=tab,
    columns={name,da,eu,fa,fi,he,id,ja,kk,ro,ur,mean},
    columns/name/.style={
      column name=Source,
      column type=l,
      string type,
    },
    columns/da/.style={/num grad={\scriptsize Danish}{\maxda}{\minda}},
    columns/eu/.style={/num grad={\scriptsize Basque}{\maxeu}{\mineu}},
    columns/fa/.style={/num grad={\scriptsize Persian}{\maxfa}{\minfa}},
    columns/fi/.style={/num grad={\scriptsize Finnish}{\maxfi}{\minfi}},
    columns/he/.style={/num grad={\scriptsize Hebrew}{\maxhe}{\minhe}},
    columns/id/.style={/num grad={\scriptsize Indonesian}{\maxid}{\minid}},
    columns/ja/.style={/num grad={\scriptsize Japanese}{\maxja}{\minja}},
    columns/kk/.style={/num grad={\scriptsize Kazakh}{\maxkk}{\minkk}},
    columns/ro/.style={/num grad={\scriptsize Romanian}{\maxro}{\minro}},
    columns/ur/.style={/num grad={\scriptsize Urdu}{\maxur}{\minur}},
    columns/mean/.style={/num grad={\scriptsize\it Mean}{\maxmean}{\minmean}},
    every head row/.style={before row=\toprule,after row=\midrule},
    every last row/.style={after row=\bottomrule},
  ]{data/clm.tsv}
}

  \caption{Cross-entropies across all source and target languages. Colors normalized by column.}
  \unskip\label{tab:clm-all}
\end{table*}

\subsection{Writing system matrix for normalized XferBench scores}
\unskip\label{sec:clm-writing-system}

See \Cref{tab:clm-writing-system,tab:clm-writing-system-type}.
Scores for reach writing system are aggregated by taking the mean.
\Cref{tab:writing-system} gives the writing system classification for the languages used in the experiments.
Although the class imbalance makes it impossible to draw any definitive claims, the preliminary results do not suggest any correlation in XferBench between the writing systems of the source and target languages.


\begin{table}
  \centering
  \begin{tabular}{lll}
    \toprule
    Type & Writing System & Language \\
    \midrule
    \multirow{4}{*}{Abjad} & \multirow{3}{*}{Arabic} & ar \\
            & & fa \\
            & & ur \\
        \cmidrule{2-3}
        & Hebrew & he \\
        \midrule
    Abugida & Devanagari & hi \\
    \midrule
    \multirow{10}{*}{Alphabet} & \multirow{2}{*}{Cyrillic} & kk \\
            & & ru \\
        \cmidrule{2-3}
        & Hangul & ko \\
        \cmidrule{2-3}
        & \multirow{7}{*}{Latin} & da \\
            & & es \\
            & & eu \\
            & & fi \\
            & & fr \\
            & & id \\
            & & ro \\
    \midrule
    Logographic & Chinese & zh \\
    \midrule
    Mixed & Japanese & ja \\
    \bottomrule
  \end{tabular}
  \caption{%
    Coarse and fine classifications of writing systems of human languages (source and target) used in the experiments.
  }
  \unskip\label{tab:writing-system}
\end{table}

\begin{table*}
  \centering
  \def\datapath{data/clm-writing-system.tsv}

\def\minval{-1.596999365469606} \def\maxval{1.9290681566044732}

{
  \pgfplotstabletypeset[
    col sep=tab,
    columns={Source,Arabic,Cyrillic,Hebrew,Japanese,Latin},
    columns/Source/.style={
      column type=l,
      string type,
    },
    columns/Arabic/.style={/num grad={Arabic}{\maxval}{\minval}},
    columns/Cyrillic/.style={/num grad={Cyrillic}{\maxval}{\minval}},
    columns/Hebrew/.style={/num grad={Hebrew}{\maxval}{\minval}},
    columns/Japanese/.style={/num grad={Japanese}{\maxval}{\minval}},
    columns/Latin/.style={/num grad={Latin}{\maxval}{\minval}},
    every head row/.style={before row=\toprule,after row=\midrule},
    every last row/.style={after row=\bottomrule},
  ]{\datapath}
}

  \caption{%
    Normalized XferBench scores by writing system (lower is better).
    Color is normalized across all values.
  }
  \unskip\label{tab:clm-writing-system}
\end{table*}

\begin{table*}
  \centering
  \def\datapath{data/clm-writing-system-type.tsv}

\def\minval{-1.4613414181868514} \def\maxval{1.4665673232827234}

{
  \pgfplotstabletypeset[
    col sep=tab,
    columns={Source,Abjad,Alphabet,Mixed},
    columns/Source/.style={
      column type=l,
      string type,
    },
    columns/Abjad/.style={/num grad={Abjad}{\maxval}{\minval}},
    columns/Alphabet/.style={/num grad={Alphabet}{\maxval}{\minval}},
    columns/Mixed/.style={/num grad={Mixed}{\maxval}{\minval}},
    every head row/.style={before row=\toprule,after row=\midrule},
    every last row/.style={after row=\bottomrule},
  ]{\datapath}
}

  \caption{%
    Normalized XferBench scores by writing system type (lower is better).
    Color is normalized across all values.
  }
  \unskip\label{tab:clm-writing-system-type}
\end{table*}

\subsection{Scatter plots for XferBench and MT}
\begin{figure*}
  \centering
  \includegraphics[width=\textwidth]{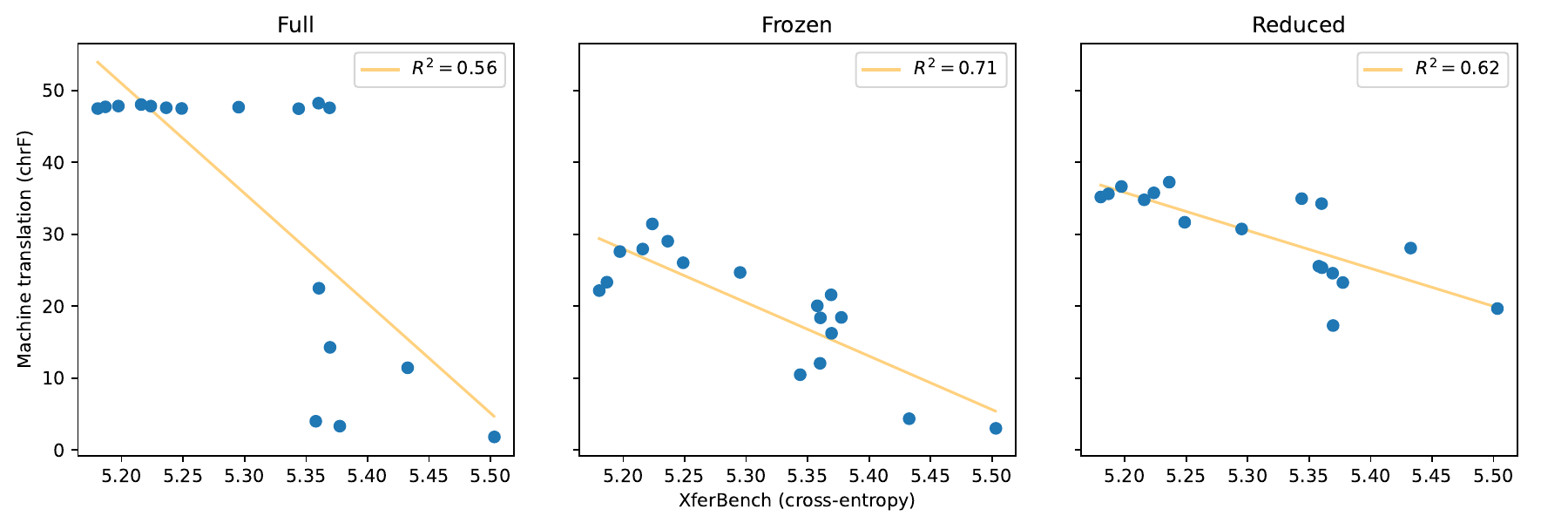}
  \caption{Scatter plots showing XferBench score versus machine translation score.}
  \unskip\label{fig:scatter}
\end{figure*}

\unskip\label{sec:scatter}
See \Cref{fig:scatter}.

\section{Cross-entropy confidence interval computation}
\unskip\label{sec:bootstrapping}

Let $s \in S$ and $t \in T$ represent source and target languages, respectively.
$h_{s,t}$ represents the test cross-entropy of a model pretrained on $s$ and evaluated on $t$.
As sated in \Cref{eq:hs}, the score on XferBench is the mean cross-entropy across all target languages:
\begin{align}
  h_{s} &= \mean_{t' \in T}\left( h_{s,t'}\right)
  .
\end{align}
We would like to calculate a confidence interval (i.e., $h^-_s$ and $h^+_s$) for a source language's mean cross-entropy using the different cross-entropies on the target languages (i.e., $h_{s,t}$ for $t \in T$), yet these samples are not i.i.d., since the mean of cross-entropy each \emph{target} language can vary.
Thus, if we would like to use bootstrapping to calculate confidence intervals, we must first normalize the cross-entropies.
Let $\hat h_{s,t}$ be the normalized score:
\begin{align}
  \hat h_{s,t} &= \frac{h_{s,t} - \mean_{s'\in S}\left(h_{s',t}\right)}{\stdev_{s'\in S}\left(h_{s',t}\right)}
  .
\end{align}
Given the normalized scores, we can now bootstrap in order to compute confidence intervals for $\hat h_s$ (i.e., in the normalized space).\footnote{This is not intended to be statistically rigorous. Our cross-entropies are unlikely to be normally distributed, but this still be helpful for generally gauging uncertainty.}
Let $\hat h^+_s$ and $\hat h^-_s$ be the upper and lower bounds of the confidence interval computed using bootstrapping in the normalized space.
We can now translate these back into the raw cross-entropy space using the means and standard deviations from before:
\begin{align}
  h^+_s &= \hat h^+_s \cdot \stdev_{s'\in S}\left(h_{s',t}\right) + \mean_{s'\in S}\left(h_{s',t}\right) \\
  h^-_s &= \hat h^-_s \cdot \stdev_{s'\in S}\left(h_{s',t}\right) + \mean_{s'\in S}\left(h_{s',t}\right)
  .
\end{align}

\section{Error analysis}
\unskip\label{sec:error-analysis}
In the \emph{Full} setting of the machine translation task, the \emph{Yao+} and \emph{Mu+, SW} settings perform worse than expected (\emph{a priori} and compared to the other results in the setting).
Validation loss converged while chrF and BLEU scores remained near zero.
We provide a couple examples (taken from the predefined test set of WMT 2014) of model output to provide some insight into the reason for this.
No post processing used, generation is capped at $50$ tokens, and ``\textbackslash{}u0000'' represent single non-printable characters.

\paragraph{Example 1}
\emph{Input}: ``And while Congress can't agree on whether to proceed, several states are not waiting.''

\emph{Reference}: ``Et tandis que les membres du Congrès n'arrivent pas à se mettre d'accord pour savoir s'il faut continuer, plusieurs États n'ont pas attendu.''

[Model pretrained on] \emph{French}:
``\#Et alors que le Congrès ne peut pas convenir de poursuivre, plusieurs États ne sont pas en attente. » (traduction libre) Le Parlement européen. Le Parlement européen est d’avis que le Parlement européen doit être en mesure de faire preuve\#''

\emph{Disc, large}:
``\#Et bien que le Congrès ne puisse pas convenir de la marche à suivre, plusieurs États ne sont pas en attente.\textbackslash{}u2028\textbackslash{}u2028[Traduit par la Rédaction]\textbackslash{}u2028(Traduit par la Rédaction)\textbackslash{}u2028(Tra\#''

\emph{Yao+}:
``\#Annexe II, par.\allowbreak{}\textbackslash{}xa02.\allowbreak{}2.\allowbreak{}2.\allowbreak{}2.\allowbreak{}2.\allowbreak{}2.\allowbreak{}2.\allowbreak{}2.\allowbreak{}2.\allowbreak{}2.\allowbreak{}2.\allowbreak{}2.\allowbreak{}2.\allowbreak{}2.\allowbreak{}2.\allowbreak{}2.\allowbreak{}2.\allowbreak{}2.\allowbreak{}2.\allowbreak{}2.\allowbreak{}2.\allowbreak{}\#''

\emph{Mu+, SW}:
``\#Annexe II.\allowbreak{}\textbackslash{}xa0\allowbreak{}\textbackslash{}xa0: Appendice I.\allowbreak{}\textbackslash{}xa0I.\allowbreak{}\textbackslash{}xa0I.\allowbreak{}\textbackslash{}xa0I.\allowbreak{}\textbackslash{}xa0I.\allowbreak{}\textbackslash{}xa0I.\allowbreak{}\textbackslash{}xa0I.\allowbreak{}\textbackslash{}xa0I.\allowbreak{}\textbackslash{}xa0I.\allowbreak{}\textbackslash{}xa0I.\allowbreak{}\textbackslash{}xa0I.\allowbreak{}\textbackslash{}xa0I.\allowbreak{}\textbackslash{}xa0I.\allowbreak{}\textbackslash{}xa0I\#''

\paragraph{Example 2}
\emph{Input}:
``This really is a must for our nation.''

\emph{Reference}:
``Cela est vraiment indispensable pour notre nation.''

\emph{French}:
``\#C'est vraiment une nécessité pour notre nation. Nous devons y parvenir. Nous devons y parvenir. Nous devons y parvenir. Nous devons y parvenir. Nous devons y parvenir. Nous devons y parvenir. Nous devons y parvenir. Nous devons y\#''

\emph{Disc, large}:
``\#C'est vraiment un devoir pour notre nation. C'est un devoir.\textbackslash{}u2028\allowbreak{}\textbackslash{}u2028(...)\allowbreak{}\textbackslash{}u2028\allowbreak{}\textbackslash{}u2028(...)\allowbreak{}\textbackslash{}u2028(...)\allowbreak{}\textbackslash{}u2028(...)\allowbreak{}\textbackslash{}u2028(...)\allowbreak{}\textbackslash{}u2028(...)\allowbreak{}\textbackslash{}u2028(...)\allowbreak{}\textbackslash{}u2028(\#''

\emph{Yao+}:
``\#Annexe II, par.\allowbreak{}\textbackslash{}xa02.\allowbreak{}2.\allowbreak{}2.\allowbreak{}2.\allowbreak{}2.\allowbreak{}2.\allowbreak{}2.\allowbreak{}2.\allowbreak{}2.\allowbreak{}2.\allowbreak{}2.\allowbreak{}2.\allowbreak{}2.\allowbreak{}2.\allowbreak{}2.\allowbreak{}2.\allowbreak{}2.\allowbreak{}2.\allowbreak{}2.\allowbreak{}2.\allowbreak{}2.\allowbreak{}\#''

\emph{Mu+, SW}:
``\#Annexe II.\allowbreak{}\textbackslash{}xa0\allowbreak{}\textbackslash{}xa0: Appendice I.\allowbreak{}\textbackslash{}xa0I.\allowbreak{}\textbackslash{}xa0I.\allowbreak{}\textbackslash{}xa0I.\allowbreak{}\textbackslash{}xa0I.\allowbreak{}\textbackslash{}xa0I.\allowbreak{}\textbackslash{}xa0I.\allowbreak{}\textbackslash{}xa0I.\allowbreak{}\textbackslash{}xa0I.\allowbreak{}\textbackslash{}xa0I.\allowbreak{}\textbackslash{}xa0I.\allowbreak{}\textbackslash{}xa0I.\allowbreak{}\textbackslash{}xa0I.\allowbreak{}\textbackslash{}xa0I\#''

\paragraph{Discussion}
Although all of the models have trouble terminating properly, the \emph{French} and \emph{Disc, large} models (which have high chrF scores) clearly condition their generation on the text, whereas \emph{Yao+} and \emph{Mu+, SW} give the same output regardless of the input.
Although this is unexpected, we can see in the \emph{Full} setting in \Cref{fig:scatter} that there is sharp drop off between high-performing and low-performing languages.
We suspect that the higher learning rate during tuning caused this bimodal distribution of results and is at least in part responsible for the poor performance \emph{Yao+} and \emph{Mu+, SW} models on the MT experiment's \emph{Full} setting.

\typeout{INFO: \arabic{comment} comments.}

\end{document}